\documentclass[10pt,twocolumn,letterpaper]{article}

\usepackage[pagenumbers]{wacv}

\newcommand{\TODO}[1]{}

\definecolor{wacvblue}{rgb}{0.21,0.49,0.74}
\usepackage[breaklinks,colorlinks,allcolors=wacvblue]{hyperref}

\def\wacvPaperID{1430}
\def\confName{WACV}
\def\confYear{2027}

\title{DeSeG: Decoupling Semantic Intent and Geometric Constraints for Physically Plausible Human-Scene Interaction}

\author{%
\begin{tabular}{c}
Jiakun Li\textsuperscript{1,\textcolor{wacvblue}{*}} \quad
Zhe Li\textsuperscript{1,\textcolor{wacvblue}{*}} \quad
Wenqiang Wu\textsuperscript{1} \quad
Zheng Chang\textsuperscript{1} \\
Mingqi Gao\textsuperscript{4} \quad
Jinyu Yang\textsuperscript{3,$\dagger$} \quad
Feng Zheng\textsuperscript{1,2,$\dagger$}
\\[0.55em]
\textsuperscript{1}Southern University of Science and Technology \quad
\textsuperscript{2}Spatialtemporal AI \\
\textsuperscript{3}Harbin Institute of Technology, Shenzhen \quad
\textsuperscript{4}University of Sheffield
\\[0.35em]
\textsuperscript{\textcolor{wacvblue}{*}}Equal contribution \quad
\textsuperscript{$\dagger$}Corresponding authors
\end{tabular}%
}

\begin{document}
\maketitle
\begin{abstract}
Synthesizing physically plausible human–scene interactions (HSI) remains a critical challenge in computer vision and the development of human avatars. Although recent generative models enable diverse motion synthesis, they suffer from an inductive bias referred to as semantic-geometric entanglement. Because spatial constraints often strongly correlate with specific actions in training data, monolithic models will learn the shortcut bias, aggressively overriding the semantic intent when faced with strict geometric~\mbox{cues}. Furthermore, this entanglement exacerbates physical hallucinations, such as body-scene penetrations. To address these limitations, we propose DeSeG, a hierarchical framework that explicitly decouples semantic intent from geometric constraints. First, we introduce a Residual Semantic Planner that encodes textual instructions and canonicalized goal voxels into a compact latent space, enabling fine-grained semantic control independent of spatial trajectories. Second, we propose a physics regularized diffusion executor that incorporates differentiable repulsive potential fields directly into the diffusion objective, enforcing collision-aware motion generation. Extensive experiments on the Lingo dataset demonstrate that DeSeG achieves state-of-the-art performance, reducing mean scene penetration by 47\% and improving semantic alignment by 29\% over the sota~\mbox{baselines}.
\end{abstract}

\begin{figure}[t]
\centering
\includegraphics[width=\linewidth]{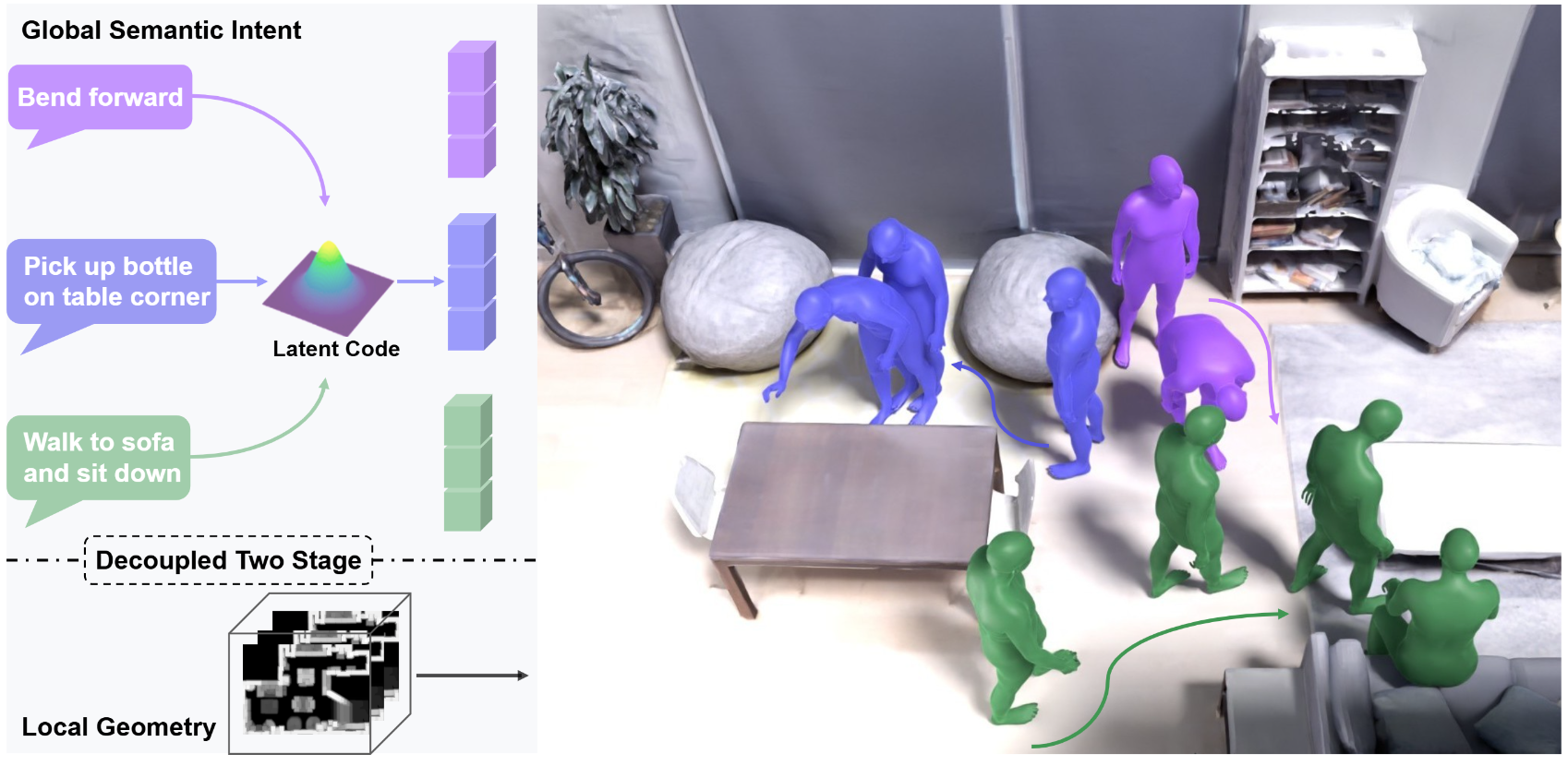}
\caption{We present \textbf{DeSeG}, a method to generate long-term physically plausible human-scene interactions by explicitly decoupling semantic intent from geometric constraints. Given a scene and a language instruction, our model first infers the affordance of interactions in latent space, and then uses physical regularization to synthesize a collision-free motion sequence, resulting in diverse and high-quality results.}
\label{fig:teaser}
\end{figure}

\section{Introduction}
\label{sec:intro}
Ensuring digital humans move in a physically consistent way within 3D scenes is crucial for immersive experiences, virtual reality, and embodied intelligence \cite{savva2019habitat, zhang2022egobody}. Recent years have seen significant progress in conditional human-scene interaction (HSI) modeling that matches high-level intent with the physical environment \cite{karunratanakul2023guided,xiao2023unified,zhao2023synthesizing, hassan2021populating}. In particular, powerful generative models such as variational autoencoders and diffusion models synthesize long-term, realistic actions from language prompts and scene context \cite{zhao2023synthesizing, guo2022generating}, marking an important step toward flexible action synthesis for human avatars in complex scenes.

Despite this progress, current methods still exhibit two critical limitations that impede robust deployment in complex and unstructured scenes:

\textbf{Semantic-Geometric Entanglement.} Existing frameworks typically conflate high-level interaction intent with low-level spatial trajectories. In diffusion-based models such as Lingo \cite{jiang2024autonomous}, the text-conditioned motion generator implicitly concatenates semantic intent to geometric targets provided as goal locations. For example, a goal on the bed often results in a lying motion. We argue that this is not merely a rare artifact, but a pervasive instance of \textit{shortcut learning} in multimodal generative models. This entanglement arises because there is no explicit mechanism to decouple `how to perform' from `where to go', leading to incorrect behaviors when geometric cues conflict with the underlying textual~\mbox{instruction}.

\textbf{Physical Hallucinations.} Generative motion models fundamentally learn data distributions without explicit physical laws, which can result in unrealistic outcomes, such as body-scene penetrations or implausible contacts. While some methods attempt to mitigate such violations using post-processing or heuristic guidance, these approaches are computationally expensive and often produce unnatural motion artifacts. Moreover, they do not guarantee consistent avoidance of collisions throughout the generation~\mbox{process}.

Based on the above insights, we introduce \textbf{DeSeG}, a hierarchical framework that explicitly decouples semantic intent from geometric feasibility for HSI synthesis. The proposed framework comprises two novel components.
First, a conditional latent module encodes textual instructions and canonicalized goal voxels into a compact intent code, so the latent space captures pure semantic affordance and conditions the subsequent motion generator.
Second, an autoregressive diffusion module embeds differentiable repulsive potential fields, derived from scene geometry, directly into the denoising objective. Penalizing predicted poses against penetration during training teaches the model to perceive and avoid obstacles intrinsically, yielding physically constrained motions without test-time~\mbox{optimization}.

By integrating decoupling mechanisms, our framework maintains the character's motion within realistic, physically plausible long-term sequences. Experiments on the Lingo and TRUMANS datasets demonstrate significant improvements in both physical plausibility and semantic consistency compared to current baselines. Our contributions are summarized as~\mbox{follows}:

\begin{itemize}
    \item We introduce a novel latent learning mechanism encoding textual instructions and local goal voxels into a latent space. This enables precise semantic control and robust generalization, even under conflicting geometric cues.

    \item We formulate a physics-regularized objective by integrating differentiable repulsive potential fields directly into the denoising loss. This endows the executor with a collision-avoidance reflex, yielding physically plausible motions without the computational overhead of test-time~\mbox{optimization}.

    \item We propose DeSeG, a hierarchical framework that explicitly decouples high-level semantic planning from low-level geometric execution. This resolves the conflict between semantic and geometric controllability in HSI~\mbox{synthesis}.
\end{itemize}

\section{Related Work}

\subsection{Human-Scene Interaction Synthesis}
Human motion synthesis has evolved from isolated character animation to modeling complex interactions within 3D environments. Early works focused on single-person motion generation conditioned on text or action labels \cite{guo2022generating, tevet2022human, chen2023mld, guo2024momask, dabral2023mofusion, cen2024generating}, using VAEs or diffusion models to produce high-fidelity motions but largely neglecting environmental context.
To address object manipulation, research expanded to Human-Object Interaction (HOI), synthesizing motions conditioned on specific objects \cite{zhang2024generating, pi2023hierarchical}. However, these approaches are typically limited to short, atomic interactions (e.g., grasping, sitting) and lack the capability to navigate complex layouts.

Recently, Human-Scene Interaction (HSI) has emerged to tackle long-term, multi-stage behaviors in large-scale scenes. HUMANISE \cite{wang2022humanise} establishes the foundational paradigm of language-conditioned HSI by aligning textual commands with captured motions in 3D scans, and Move-as-You-Say \cite{wang2024move} further grounds language-guided motion in scene affordances. PROX \cite{hassan2019resolving} enforces scene-aware static pose optimization, while SceneDiffuser \cite{huang2023scenediffuser} and Lingo \cite{jiang2024autonomous} formulate interaction synthesis as scene-conditioned diffusion, enabling multi-stage tasks in cluttered environments. More recent efforts pursue hierarchical or geometry-grounded decompositions: SceMoS \cite{ghosh2026scemos} disentangles a text-conditioned global planner from local execution using lightweight 2D scene cues (bird's-eye-view maps and local heightmaps), reporting state-of-the-art realism on TRUMANS, while HOSIG \cite{yao2026hosig} couples a scene-aware grasp generator, a collision-aware navigation planner, and a trajectory-controlled synthesizer to compose long-horizon full-body human--object--scene interactions.
Despite this progress, current methods rely on goal locations as strong conditioning, and even multi-stage pipelines such as Lingo couple textual intent and geometric targets in one backbone; when goal and instruction conflict, the geometric prior dominates. We instead introduce a hierarchical decomposition of high-level semantic planning and low-level geometric execution, retaining the goal but structurally separating its influence.

\subsection{Latent Representations for Human Motion}
Latent representations serve as the bridge between high-level semantics and low-level kinematics. Previous approaches leverage VAE, VQ-VAE, and transformer-based architectures \cite{petrovich2022temos, guo2022tm2t, zhang2023t2m} to compress motion into compact latent codes. To achieve semantic control, MotionCLIP \cite{tevet2022motionclip} aligns the motion latent space with the CLIP text embedding, enabling zero-shot generation.
Most latent spaces target free-space motion and ignore spatial constraints; scene-conditioned models inject goal voxels directly into the diffusion model, entangling instruction and geometry in one backbone. We instead introduce a Residual Semantic Planner that produces a compact latent code before geometric diffusion, preventing spatial signals from overwhelming semantic intent.

The most closely related effort is SemGeoMo \cite{cong2025semgeomo}, which likewise pursues a two-stage decoupling for HOI. The key distinction lies in \emph{how} the two factors are separated. SemGeoMo predicts an \emph{explicit} intermediate contact geometry that guides the subsequent motion, which is informative but tightly tied to a predefined contact parameterization and offers no built-in penetration guarantee. In contrast, DeSeG learns an \emph{implicit} disentangled affordance latent through a residual CVAE and, crucially, internalizes physical plausibility directly into the diffusion training objective via a differentiable repulsive potential field. This removes the need for explicit contact annotation and for test-time optimization, and, unlike prior decoupling work, we explicitly stress-test the separation under conflicting semantic--geometric cues through our NC-Bench.

\subsection{Physics-Aware Human Motion Synthesis}
For physical plausibility, one paradigm uses physics simulators \cite{yuan2023physdiff} or reinforcement learning \cite{peng2021amp, xiao2023unified}, but simulators are often non-differentiable and hard to integrate into end-to-end generative~\mbox{pipelines}.
Another prevailing approach is Test-Time Guidance \cite{karunratanakul2023guided, zhang2020place, zhao2023synthesizing}, where SDF-based collision losses or contact optimizations are used to guide the sampling process during inference. While this improves quality, it significantly increases computational latency, hindering real-time applications.
To achieve both physical plausibility and inference efficiency, recent works integrate geometric losses directly into training. For instance, SceneDiffuser \cite{huang2023scenediffuser} applies scene-penetration penalties during~\mbox{optimization}.
In contrast, we embed the Potential Field \cite{xue2026humanoid} into the diffusion training objective, so the model intrinsically generates collision-free trajectories without expensive test-time guidance, balancing physical constraints with~\mbox{diversity}.

\begin{figure*}[t]
\centering
\includegraphics[width=0.85\textwidth]{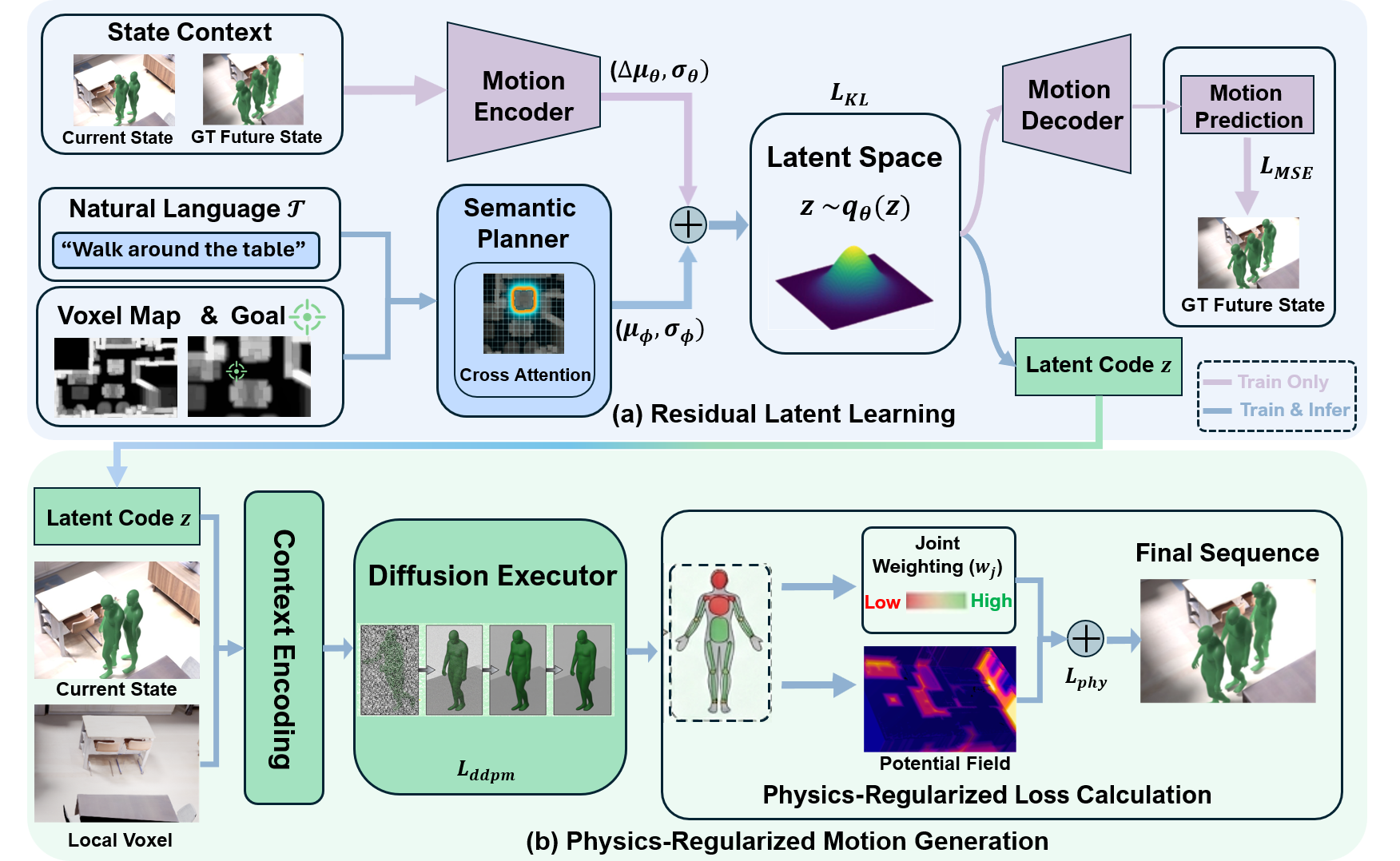}
\caption{\textbf{Overview of DeSeG framework.} Our method decouples HSI synthesis into two stages. \textbf{Stage 1: Residual Latent Learning.} Learning a compact latent affordance code by residual CVAE. To prevent spatial entanglement, the goal voxel is canonicalized before interacting with the textual instruction via cross-attention.  \textbf{Stage 2: Physics-Regularized Motion Generation.} The Motion Diffusion Executor predicts the future state conditioned on the extracted latent code, the current state, and the local scene voxel. The gradients of $L_{Phys}$ are backpropagated alongside the diffusion objective $L_{ddpm}$, instilling collision-avoidance into the generation process.}
\label{fig:model}
\end{figure*}

\section{Method}
\label{sec:method}

In this section, we present the mathematical formulation and detailed architectural implementation of \textbf{DeSeG}. As illustrated in \cref{fig:model}, Our method factorizes the complex human motion generation into two processes: a \textbf{Residual Semantic Planner} that infers a latent interaction affordance via conditional multimodal inputs, and a \textbf{Physics-Regularized Diffusion Executor} that grounds latent code into physically plausible motion sequences through a potential field diffusion training strategy.

\subsection{Problem Formulation}

We define the HSI synthesis task as learning a conditional generative model:

\begin{equation}
p(\mathbf{M} \mid \mathcal{T}, \mathcal{S}, \mathbf{g}).
\end{equation}

Given a 3D scene $\mathcal{S}$ and a target interaction position or object $\mathbf{g} \in \mathbb{R}^3$, our goal is to generate a plausible human-scene interaction, where the motion style can be controlled by a user-specified natural language instruction
$\mathcal{T} = \{w_1, \dots, w_L\}$. Instructions are provided at the action-segment level, \ie a single command describes one interaction segment, and a long-horizon sequence is composed of consecutive segments.
The output is a motion sequence $\mathbf{M} \in \mathbb{R}^{T \times J \times D}$ over $T$ timesteps, where the model operates directly in 3D joint-position space with $J{=}28$ body joints and $D{=}3$ (in a $y$-up frame). The corresponding SMPL-X \cite{loper2015smpl, pavlakos2019smplx} body parameters are recovered from the predicted joints through a lightweight learned joints-to-SMPL-X regressor, rather than a classical inverse-kinematics~\mbox{optimization}.

Previous diffusion-based approaches model this distribution with a monolithic backbone $\epsilon_\varphi(\mathbf{M}_t, t, \mathcal{C})$, where $\mathcal{C}$ concatenates conditioning signals such as text and geometry. This architecture suffers from a severe inductive bias: in typical HSI datasets the geometric goal $\mathbf{g}$ is a dense, high-frequency signal highly predictive of the action category, so optimization latches onto $\mathbf{g}$ while ignoring the subtle linguistic signal $\mathcal{T}$, causing failures on negative constraints.

This is an \emph{architectural} pathology that loss-level regularization cannot remove: sharing one conditioning pathway lets the gradient of $\mathbf{g}$ structurally dominate that of $\mathcal{T}$. Our decomposition instead imposes an information bottleneck---routing semantics through a compact latent $\mathbf{z}$ computed from a \emph{canonicalized}, trajectory-free view of the geometry prevents absolute spatial configuration from leaking into the semantic factor. We test this decoupling advantage directly in our ablations (\cref{tab:quant_comparison}).

We therefore propose a two-stage decoupled formulation via a latent code $\mathbf{z}$ that encodes interaction affordance from multimodal inputs, emphasizing the semantic ``what'' and ``how''. We decompose the joint distribution as:

\begin{equation}
\resizebox{0.85\columnwidth}{!}{$\displaystyle
p(\mathbf{M} \mid \mathcal{T}, \mathcal{S}, \mathbf{g})
=
\int
p_\varphi(\mathbf{M} \mid \mathbf{z}, \mathcal{S}_{local}, \mathbf{g})
\cdot
p_\phi(\mathbf{z} \mid \mathcal{T}, \Psi(\mathcal{S}, \mathbf{g}))
\, d\mathbf{z}.
$}
\end{equation}

Here $\Psi(\cdot)$ is the Canonicalization Operator, and $p_\phi$, $p_\varphi$ are our \textbf{Planner} and \textbf{Executor}. This structure forces $\mathbf{z}$ to capture the interaction~\mbox{affordance}.

\subsection{Residual Semantic Planner}

Inspired by hierarchical vision-language-action methods \cite{xue2025leverb}, the Semantic Planner is the high-level reasoning module. It maps the linguistic instruction and scene geometry into a compact latent space using a residual Conditional Variational Autoencoder (CVAE) \cite{kingma2013auto, sohn2015learning}, focusing on semantic intent while offloading trajectory-specific details to the motion encoder. The sampled latent code conditions a decoder that predicts future poses $\hat{S}_{t+1}, \dots, \hat{S}_{t+M}$ from the current state $S_t$.

\paragraph{Kinematics Encoder.}
Since the Semantic Planner only observes the current conditional inputs during inference, we introduce a kinematics encoder during training to capture additional temporal information from future states. The encoder is a Multi-Layer Perceptron (MLP) that takes the flattened ground-truth future states $\mathbf{S}_{t+1:t+M}$ as input and predicts the residual mean $\Delta\mu$ and the posterior standard deviation $\sigma_{post}$ of the latent distribution.

\paragraph{Canonicalized Geometric Perception.}
We represent the scene as a full 3D occupancy grid at $2\,\mathrm{cm}$ voxel resolution for both training and inference. 
To prevent absolute-orientation leakage into the latent space, we apply the canonicalization operator $\Psi$: we crop a local voxel grid centered at the goal $\mathbf{g}$ and express it in an egocentric canonical frame via the rigid transform $R^{-1}$ that aligns the agent's heading to a fixed axis---the same frame in which motion is generated---so the goal voxel no longer depends on absolute scene orientation. The resulting $V_{can} = \Psi(\mathcal{S}, \mathbf{g})$ exposes the intrinsic shape of the interaction region (\eg seat height, table width) in an orientation-invariant manner.

We then align language with concrete geometry via cross-attention. The instruction $\mathcal{T}$ is encoded by a frozen CLIP model \cite{radford2021learning} into text tokens and $V_{can}$ by a 3D-CNN \cite{maturana2015voxnet} into geometric tokens; the text embeddings act as queries that selectively attend to the voxel features, grounding prompts to task-relevant geometric affordances. The fused features are pooled and projected through MLPs to parameterize the prior mean $\mu_\phi$ and standard deviation $\sigma_\phi$.

\paragraph{Residual Latent Space.}

To effectively bridge high-level semantic planning and low-level kinematic execution, we formulate the latent space using a residual connection. Specifically, we define the generative prior distribution as

\begin{equation}
p_\phi(z_t \mid \mathcal{T}, V_{can}) = \mathcal{N}(\mu_\phi, \sigma_\phi^2).
\end{equation}

During training, the posterior $q_\theta(z_t)$ combines the kinematics encoder and the prior network. Rather than predicting the absolute mean, the kinematics encoder predicts a residual shift $\Delta \mu_\theta$, so the posterior mean is a residual addition while the variance is taken from the encoder ($\sigma = \sigma_\theta$).

This residual design explicitly forces the kinematics encoder to capture only the fine-grained residual details that are not already inferable from the scene-language inputs, thereby allowing the semantic planner to focus entirely on the abstract intent. The network is optimized using the Evidence Lower Bound (ELBO), where a KL divergence loss $D_{KL}(q_\theta \parallel p_\phi)$ is applied to regularize the posterior towards the prior.

\subsection{Physics-Regularized Diffusion Executor}

The Executor translates the abstract semantic latent $\mathbf{z}$ into a dense, physically valid motion sequence $\mathbf{M}$ conditioned on the local scene geometry. We employ a Transformer-based Diffusion Model (DiT) \cite{ho2020denoising, peebles2023scalable} and introduce a training-time physics-aware strategy to address the physical hallucination problem common in probabilistic generative models.

\paragraph{Conditional Diffusion Backbone.}
We model the motion generation as a reverse diffusion process $p_\varphi(\mathbf{M}_{t-1} \mid \mathbf{M}_t, C)$. To bridge the gap between high-level intent and low-level execution, the core conditioning context $C$ comprises two primary components:
(1) The semantic affordance latent $\mathbf{z}$ sampled from the Planner;
(2) The local scene voxel $V_{local}$ , providing regional obstacle awareness.

To re-introduce the spatial guidance abstracted away by the Planner, we run A* \cite{hart1968formal} on the occupancy map to produce collision-free waypoints toward the goal; this path acts as a macro-level directional prompt that guides the autoregressive diffusion toward the target.

\paragraph{Differentiable Repulsive Potential Field.}
To enforce physical plausibility, we internalize a collision-avoidance mechanism inspired by the Humanoid Potential Field \cite{xue2026humanoid} directly into the diffusion training objective. Since the network predicts noise $\epsilon_\varphi$, we employ Tweedie's formula \cite{efron2011tweedie, chung2022diffusion} to analytically estimate the clean motion $\hat{\mathbf{M}}_0$ at any timestep $t$:
\begin{equation}
\hat{\mathbf{M}}_0 = \frac{\mathbf{M}_t - \sqrt{1-\bar{\alpha}_t} \cdot \epsilon_\varphi(\mathbf{M}_t, t, C)}{\sqrt{\bar{\alpha}_t}}.
\end{equation}

We extract the absolute world coordinates $o_{j,k} \in \mathbb{R}^3$ for each joint from $\hat{\mathbf{M}}_0$. To efficiently obtain the shortest Euclidean distance to environmental obstacles, we query a pre-computed high-resolution Signed Distance Field (SDF) \cite{park2019deepsdf}, denoted as $\mathcal{D}(o)$, and explicitly define a pure repulsive potential energy function $U_{rep}$ for any spatial point $o$:
\begin{equation}
U_{rep}(o) = \frac{1}{2} \eta \cdot \text{ReLU}\big(d_{safe} - \mathcal{D}(o)\big)^2,
\end{equation}
where $d_{safe}$ is a pre-defined safe margin, and $\eta$ is a scaling factor controlling the stiffness of the potential field. The ReLU activation strictly ensures that the penalty is only triggered when a joint breaches the safety threshold.

Building upon this continuous potential function, we formulate the overall physical regularization loss across all frames and skeletal joints:
\begin{equation}
\mathcal{L}_{phy}(\hat{\mathbf{M}}_0) = \sum_{k=1}^T \sum_{j=1}^J w_j \cdot U_{rep}(o_{j,k}).
\end{equation}

Treating all joints uniformly can trap humanoids in local minima, as symmetric penalties on opposing limbs~\mbox{cancel}. We therefore use joint-specific weights $w_j$: higher penalties on the torso and head prevent catastrophic penetrations, while contact-prone extremities get softer margins. Since $\hat{\mathbf{M}}_0$ is differentiable in the weights $\varphi$, the field's spatial derivative acts as a gradient-based repulsive signal; backpropagating it teaches the network to predict noise that steers kinematics away from body-scene~\mbox{penetrations}.

\paragraph{Optimization Strategy.}
We adopt a decoupled two-stage training paradigm to preserve semantic disentanglement. First, the Semantic Planner is trained to convergence using the ELBO objective to establish a stable latent affordance space. Second, the Planner's weights are frozen, and the Diffusion Executor is trained conditioned on the sampled $z \sim p_\phi(\cdot)$.

To simultaneously capture the data distribution and adhere to physical constraints, the Executor is optimized using a joint training objective:
\begin{equation}
\mathcal{L}_{overall} = \mathcal{L}_{ddpm} + \lambda \cdot \mathcal{L}_{phy}(\hat{\mathbf{M}}_0),
\end{equation}
where $\lambda$ balances the two terms. To keep the physics penalty from destabilizing training or collapsing diversity, we (i) warm up $\lambda$ from 0 to $\lambda_{\max}$ over the first 20 epochs; (ii) apply the penalty only at small timesteps $t \le 5$, where the Tweedie estimate $\hat{\mathbf{M}}_0$ is close to the clean motion and yields reliable gradients, whereas at larger $t$ it is dominated by noise and would inject instability; and (iii) use the margin $d_{\text{safe}}$ and per-joint weights $w_j$ to avoid over-penalizing fingertips. The Executor thus satisfies both constraints: it executes the behavior commanded by $\mathbf{z}$ while carving out collision-free, physically plausible trajectories.

\section{Experiments}

To rigorously evaluate the efficacy of DeSeG, we conduct extensive experiments to assess whether the proposed hierarchical framework improves semantic grounding and whether physics regularization eliminates physical inconsistency. All analyses are specifically designed to validate the structural disentanglement assumptions introduced in \cref{sec:method}.

\begin{table*}[t]
  \centering
  \small
  \renewcommand{\arraystretch}{1.05}
  \setlength{\tabcolsep}{8pt}
  \caption{\textbf{Quantitative results on Lingo and TRUMANS datasets} \cite{jiang2024autonomous}. We compare against state-of-the-art baselines (top), ablate each DeSeG component (middle), and report cross-dataset generalization to TRUMANS \cite{jiang2024scaling} scenes (bottom). All DeSeG models share the same training configuration; penetration is measured per dataset and is not comparable across datasets. Subscripts denote standard deviation over 3 random seeds; all primary-metric gaps between DeSeG and the strongest baseline exceed $2\sigma$.}
  \label{tab:quant_comparison}
  \begin{tabular}{l|cccccc}
    \toprule
    \textbf{Method} &
    \textbf{FID$\downarrow$} &
    \textbf{Diversity$\rightarrow$} &
    \textbf{M-modality$\rightarrow$} &
    \textbf{R-Precision$\uparrow$} &
    \textbf{Pene$_{\text{mean}}$$\downarrow$} &
    \textbf{Pene$_{\text{max}}$$\downarrow$} \\
    \midrule
    \multicolumn{7}{l}{\textit{Comparison with baselines}} \\
    TRUMANS \cite{jiang2024scaling} & 3.286 & 4.984 & 3.036 & 0.413 & 0.404 & 1.825 \\
    TeSMo \cite{yi2024generating} & 3.558 & 5.328 & 2.956 & 0.392 & 0.278 & \textbf{0.915} \\
    Lingo \cite{jiang2024autonomous} & 3.093 & 5.832 & \textbf{2.239} & 0.437 & 0.392 & 1.229 \\
    \midrule
    \multicolumn{7}{l}{\textit{Ablation of DeSeG components}} \\
    w/o Hierarchical Latent & 3.279 & 5.461 & 2.439 & 0.430 & 0.241 & 0.938 \\
    w/o Cross-Attention & 3.113 & 5.636 & 2.721 & 0.457 & 0.239 & 0.946 \\
    w/o Physical Regularization & 2.983 & 5.822 & 2.524 & 0.514 & 0.367 & 1.393 \\
    \midrule
    \textbf{DeSeG (full)} & \textbf{2.882} & \textbf{5.865} & 2.541 & \textbf{0.562} & \textbf{0.207} & 0.921 \\
    \midrule
    \textit{Comparison on TRUMANS} & \textbf{FID$\downarrow$} & \textbf{Diversity$\rightarrow$} & \textbf{PFC$\downarrow$} & \textbf{Cont$\uparrow$} & \textbf{Pene$_{\text{mean}}$$\downarrow$} & \textbf{Pene$_{\text{max}}$$\downarrow$} \\
    \midrule
    Lingo \cite{jiang2024autonomous} & 0.493 & 2.932 & 0.657 & 0.941 & 2.279 & 13.376 \\
    DeSeG & 0.338 & 2.853 & 0.621 & 0.986 & 1.183 & 9.187 \\
    \bottomrule
  \end{tabular}
\end{table*}

\begin{figure*}[t]
    \centering
    \makebox[0.33\linewidth]{\footnotesize\textbf{DeSeG}}%
    \makebox[0.33\linewidth]{\footnotesize\textbf{Lingo}}%
    \makebox[0.33\linewidth]{\footnotesize\textbf{TRUMANS}}\\[1pt]
    \includegraphics[width=0.95\textwidth]{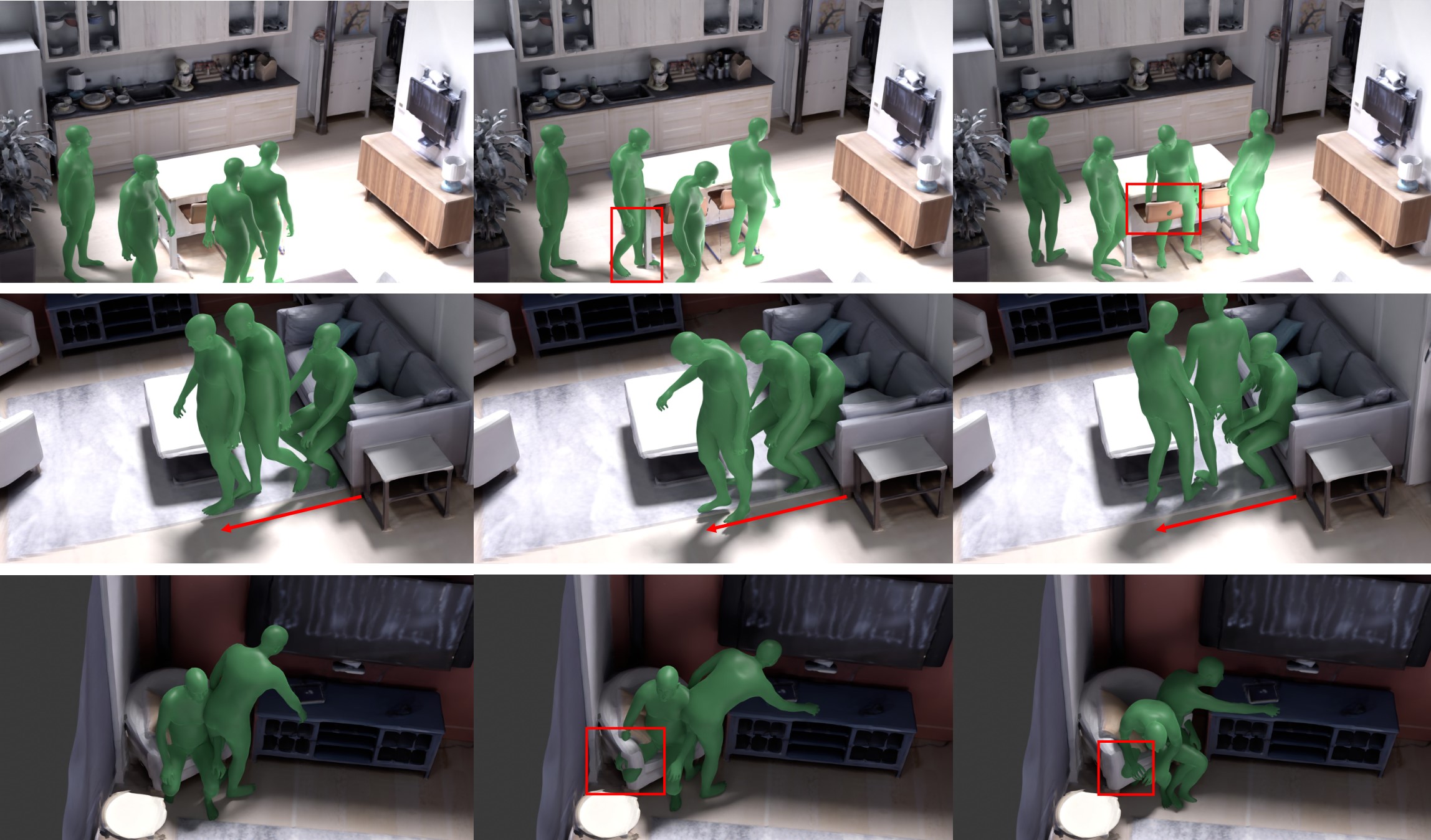}
    \caption{Qualitative comparison of 3D human--scene interaction in indoor scenes.
    \textbf{Rows} (top to bottom): \emph{``walk around the table''}, \emph{``stand up and creep''}, \emph{``pick up the book on cabinet''}.
    \textbf{Columns}: DeSeG, Lingo \cite{jiang2024autonomous}, TRUMANS \cite{jiang2024scaling}. Red boxes mark scene penetration; red curves show temporal trajectories. DeSeG gives the most reasonable result in all three cases.}
    \label{fig:visualization}
\end{figure*}

\subsection{Experimental Setup}

\paragraph{Dataset and Preprocessing.}
We train and evaluate on the large-scale Lingo dataset, which provides 16 hours of high-fidelity, text-annotated HSI motion capture across 120 indoor scenes. Following \cite{jiang2024autonomous}, we use a 4:1 scene-based train/eval split, ensuring evaluation in geometrically unseen environments. For physical regularization, we precompute high-resolution SDFs \cite{oleynikova2016signed} at $2\,\mathrm{cm}$ voxel resolution for all training scenes; SDF values are normalized per scene and joint positions are trilinearly interpolated from the grid to compute $\mathcal{D}(\mathbf{x})$. Full preprocessing details are in the appendix.

\paragraph{Evaluation Metrics.}
To provide a comprehensive assessment, we align our evaluation protocol with existing state-of-the-art HSI frameworks \cite{tevet2022human, jiang2024scaling}. For generation quality and semantic alignment, we report the Fréchet Inception Distance (FID), Diversity, Multi-modality (M-modality) and R-Precision \cite{tevet2022human}. These metrics quantify how accurately the generated motion distribution matches the linguistic intent and motion quality. For physical plausibility, we employ standard scene penetration metrics, including $Pene_{mean}$, and $Pene_{max}$.

\paragraph{Baseline Setup.}
We compare against three HSI models: Lingo~\cite{jiang2024autonomous}, TRUMANS~\cite{jiang2024scaling}, and TeSMo~\cite{yi2024generating}, all retrained on the Lingo dataset. All models use a 3D occupancy grid except TeSMo, which uses a 2D floor map. For fairness, every model is trained on identical data splits on a single NVIDIA RTX A6000 GPU from its official public configuration, with no extra tuning advantage given to our method; full protocols are in the appendix. All quantitative results are averaged over 3 random seeds (mean and standard~\mbox{deviation}).

\paragraph{Implementation Details.}
The planner pairs a frozen CLIP ViT-L/14 encoder with a 3D-CNN scene encoder (16-head cross-attention, 256-d latent); the executor is an 8-block DiT. We train the planner (50K iters), freeze it, then the executor (100K iters, batch 128, AdamW, lr $1\!\times\!10^{-4}$); the physics loss uses $d_{safe}=0.05$\,m, $\eta=0.5$. Full architecture and per-joint weights are in the appendix.

\paragraph{Negative Constraint Benchmark.}
Standard metrics often miss semantic-geometric entanglement, as test distributions mirror training biases. We therefore introduce NC-Bench, 30 scenarios in which the instruction semantically contradicts the target object's strong geometric prior; we contextualize standalone AMASS~\cite{mahmood2019amass} motions within unseen TRUMANS~\cite{jiang2024scaling} scenes. Performance is Semantic-Geometric Consistency (SGC), from a perceptual user study: 10 annotators give a binary judgment of whether the character performs the specified action without being captured by the object's canonical geometry, averaged into SGC. We use a perceptual protocol because NC-Bench targets the conflict cases that automated, distribution-matched metrics (\eg FID) systematically mis-score, inheriting the same shortcut bias as the models they assess. Inter-annotator agreement is high (Fleiss' $\kappa=0.818$); the study collects only anonymous binary judgments on synthetic clips.

\begin{table}[b]
  \centering
  \small
  \renewcommand{\arraystretch}{1.05}
  \caption{Evaluation on NC-Bench: traditional metrics (FID, R-Precision) reported alongside SGC. Standard metrics mask the entanglement that SGC exposes.}
  \label{tab:nc_bench}
  \begin{tabular}{l|ccc}
    \toprule
    \textbf{Method} & \textbf{FID$\downarrow$} & \textbf{R-Precision$\uparrow$} & \textbf{SGC\,(\%)$\uparrow$} \\
    \midrule
    TRUMANS \cite{jiang2024scaling} & 4.102 & 0.427 & 38.6 \\
    TeSMo \cite{yi2024generating} & 3.845 & 0.393 & 41.4 \\
    Lingo \cite{jiang2024autonomous} & \textbf{3.120} & 0.459 & 58.2 \\
    \textbf{Ours} & 3.155 & \textbf{0.584} & \textbf{72.3} \\
    \bottomrule
  \end{tabular}
\end{table}

\subsection{Quantitative Results}

\paragraph{Semantic Alignment and Motion Quality.}
\cref{tab:quant_comparison} compares \textbf{DeSeG} with three baselines. DeSeG attains the best FID (2.882), indicating that the Semantic Planner suppresses geometrically induced shortcuts, and the highest Diversity (5.865)---unlike constrained methods that sacrifice richness for compliance. Its Multi-modality (2.541) stays comparable to the baselines (\eg Lingo at 2.239), indicating that decoupling and physical regularization preserve per-prompt diversity rather than collapsing generation onto a single dominant mode.

\paragraph{Physical Plausibility.}
DeSeG substantially reduces scene penetration, lowering mean penetration from 0.392 (Lingo) to 0.207 and maximum penetration from 1.229 to 0.921. It also matches TeSMo, which applies SDF-based guidance during sampling, despite a fundamental difference: TeSMo enforces physics through post-hoc test-time optimization, whereas DeSeG internalizes collision awareness directly into the diffusion objective. The suppression of catastrophic penetrations indicates that the backbone has learned a robust collision-aware prior.

\paragraph{Inference Efficiency.}
Internalizing physics during training removes any test-time SDF optimization or post-processing, yielding roughly a $20\%$ inference speed-up over TeSMo while attaining lower penetration. The autoregressive executor also has a fixed per-segment cost, so runtime grows linearly with the number of action segments.

\paragraph{Cross-Dataset Generalization.}
A central concern of HSI (Human-Scene Interaction) methods is their generalization capability across different training distributions. We retrain our model on TRUMANS~\cite{jiang2024scaling} dataset, also transferring the Lingo baseline identically and following the TRUMANS protocol---reporting foot-sliding (PFC$\downarrow$) and contact (Cont$\uparrow$) in place of M-modality and R-Precision (\cref{tab:quant_comparison}, bottom). Despite the domain shift, DeSeG transfers more robustly---lower FID ($0.338$ vs.\ $0.493$) and roughly half the penetration ($1.183$ vs.\ $2.279$)---showing the decoupled latent and physics-regularized executor are effective for data with various distributions.
\paragraph{Results on NC-Bench.}
As shown in \cref{tab:nc_bench}, FID masks the limitations of existing methods out of distribution: Lingo keeps a competitive FID (3.120) yet its Semantic-Geometric Consistency (SGC) drops sharply. This confirms that previous methods suffer from severe entanglement, triggering canonical interactions dictated by the target position while overriding the conflicting instruction.

In contrast, by decoupling semantic intent from geometric priors, DeSeG reaches an SGC of 72.3\%, exceeding the strongest baseline by over 14\% absolute without sacrificing motion quality (comparable FID and R-Precision)---validating that it generates semantically faithful, physically plausible interactions even when instructions diverge from default affordances.

\subsection{Ablation Studies}

To understand the contribution of each component of DeSeG, we perform an ablation study by removing key modules and observing the impact on performance in \cref{tab:quant_comparison}. Different variants were trained on the same dataset using a single NVIDIA A6000 GPU.

\paragraph{Ablating the Semantic Planner.}
\textbf{w/o Hierarchical Latent:} We remove the CVAE Semantic Planner and inject the raw CLIP text embeddings and global scene voxels directly into the Diffusion Executor. This reduces our framework to a strong monolithic architecture that keeps the same backbone, conditioning, and physics regularization---only the decoupling is removed---\ie the controlled monolithic baseline called for in \cref{sec:method}. R-Precision drops to 0.430 and FID to 3.279: without $\mathbf{z}$, dense spatial tokens overwhelm the sparse textual signals and the network regresses to shortcut behavior, confirming that loss-level conditioning over a shared pathway does not overcome the entanglement.
\textbf{w/o Cross-Attention:} Replacing geometry-aware cross-attention with simple feature concatenation drops R-Precision to 0.457 and raises FID to 3.113, showing that explicit attention routing is necessary to ground language into spatial affordances.

\paragraph{Ablating the Physics-Regularized Executor.}
\textbf{w/o Physical Regularization:} Removing $\mathcal{L}_{phy}$ from the training objective abolishes physical awareness: $Pene_{mean}$ rises to 0.367 and $Pene_{max}$ to 1.393. The unregularized model generates semantically correct yet physically~\mbox{implausible motions}.

\begin{figure}[t]
\centering
\includegraphics[width=\linewidth]{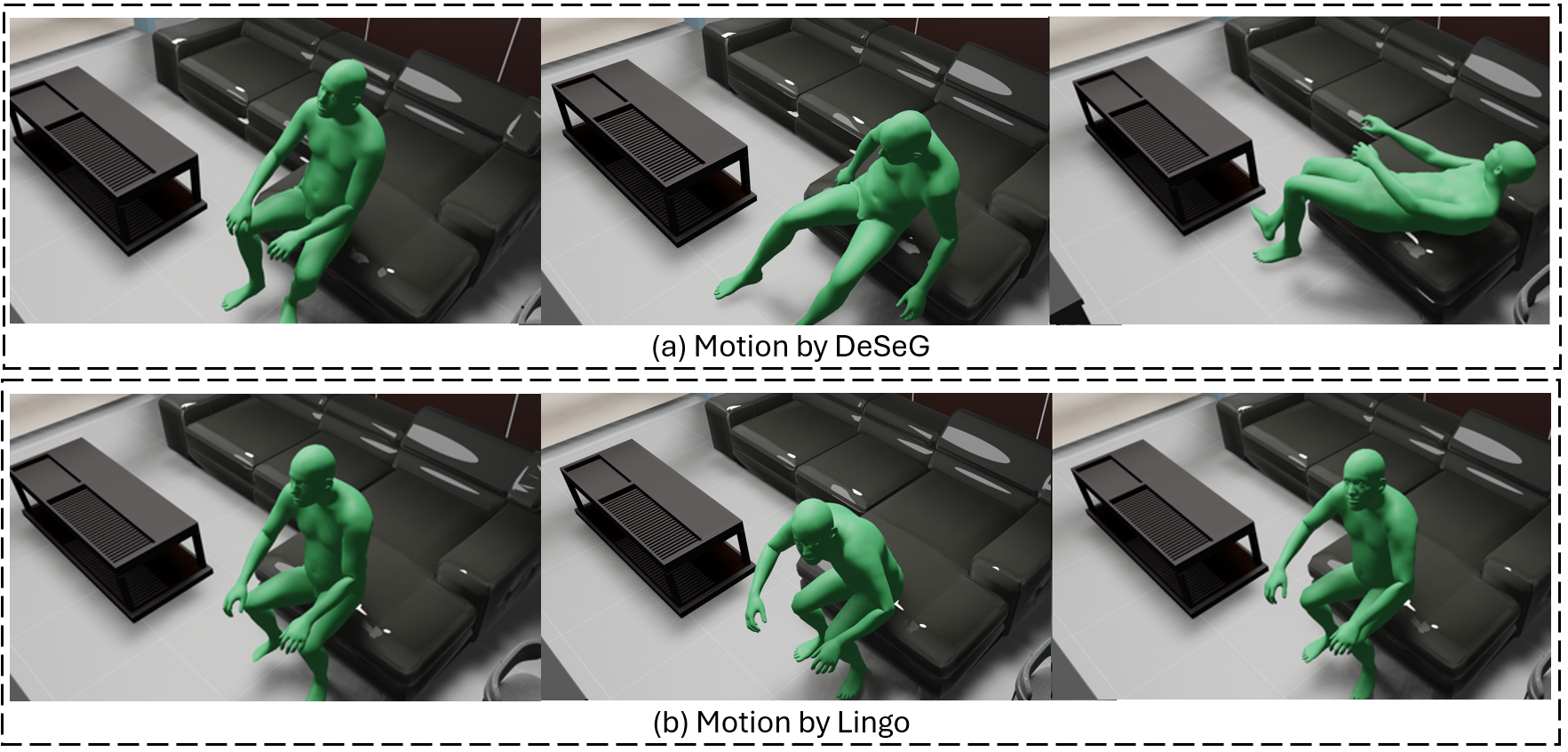}
\caption{Visualization in a TRUMANS \cite{jiang2024scaling} scenario for the instruction \textbf{sit down on sofa and then sleep}: Lingo freezes in a sitting posture, whereas DeSeG transitions into a lying pose.}
\label{fig:vis}
\end{figure}

\subsection{Qualitative Analysis and Limitations}

\paragraph{Visualization.}
\cref{fig:visualization} compares DeSeG with baselines on three multi-action instructions across Lingo and TRUMANS scenes: our avatars navigate around obstacles and rest hands on armrests without mesh penetration (red cues), showing that the executor internalizes the humanoid potential field. \cref{fig:vis} further shows a sofa$\rightarrow$sleep case in which Lingo freezes in a sitting posture---the sofa's geometric prior overwhelms the later textual condition---whereas DeSeG transitions smoothly into a lying pose, confirming~\mbox{successful decoupling}.

\paragraph{Limitations.}
Failure cases occur in cluttered environments or with geometrically abstract objects lacking clear affordances, exposing the limits of voxel-based geometry encoding; finer-grained representations are a natural next step. A second limitation is the decoupling trade-off: for affordance-critical tasks demanding tight semantic--geometric coupling (\eg facing the screen when ``watching television''), routing intent through a compact, trajectory-free latent may discard fine-grained alignment cues---modulating the decoupling strength per interaction type is a promising direction.

\section{Conclusion}
We presented DeSeG, a hierarchical framework that resolves semantic-geometric entanglement and physical hallucination in HSI. By structurally decoupling generation, the Residual Semantic Planner extracts interaction affordances via latent codes, while the Physics-Regularized Diffusion Executor gains an intrinsic collision-avoidance reflex without test-time optimization. Experiments show state-of-the-art quality and physical plausibility, especially in negative-constraint scenarios where monolithic~baselines~fail.

\clearpage
\appendix
\section{Additional Details}
\label{app:additional}

This appendix provides details omitted from the main text due to space
constraints. Section~\ref{app:impl} gives implementation details (data
preprocessing, network architectures, optimization). Section~\ref{app:addexp}
reports an A* path-guidance ablation and statistical-significance details, and
Section~\ref{app:ncbench} lists the Negative Constraint Benchmark scenarios.

\section{Implementation Details}
\label{app:impl}

\subsection{Data Preprocessing and SDF Generation}
We precompute a 3D Signed Distance Field (SDF) for each Lingo scene from a 
3D occupancy grid at $2\,\mathrm{cm}$ resolution. Voxel grids from
Blender meshes often have hollow interiors; rather than closing them with a 2.5D
elevation filling---which marks free space \emph{under} furniture (\eg{} beneath a
table) as solid, inconsistent with under-object interactions such as NC-Bench cases
14--17---we resolve the interiors directly, so only genuine surfaces and solids are
occupied while traversable space beneath objects is preserved. The signed grid is
obtained via \texttt{distance\_transform\_edt()} (CuPy) on the occupied and free
regions; the repulsive field then reads SDF values at each joint position.

\subsection{Network Architecture Details}
\textbf{Residual Semantic Planner:} A frozen CLIP ViT-L/14 text encoder yields an
$L_2$-normalized $768$-d embedding, concatenated with pelvis/hand goal embeddings and
fused by a linear layer to form the attention query. The local goal-centric scene
voxel is encoded by a $2$-layer 3D-CNN, flattened, and projected to the model
dimension. A $16$-head cross-attention uses the fused text-goal embedding as Queries
and the 3D-CNN features as Keys/Values. The prior network, kinematics encoder, and
motion decoder are MLPs with hidden dimension $[512,512]$; the affordance latent $z$
has dimension $256$.

\textbf{Physics-Regularized Diffusion Executor:} A DiT backbone of $8$ Transformer
blocks ($16$ heads, hidden dimension $512$). The latent $z$ is injected into each
block via MLP layers, and the local scene voxel is encoded by a lightweight ViT and
injected via cross-attention.

\subsection{Optimization and Physics Hyperparameters}
The diffusion network is optimized using the AdamW optimizer with a learning rate
of $1 \times 10^{-4}$ and a cosine annealing schedule. We adopt a decoupled
two-stage training paradigm. The Semantic Planner is trained for $50$K iterations.
Subsequently, its weights are frozen, and the Diffusion Executor is trained for
$100$K iterations with a batch size of $128$. For the physical regularization loss
$\mathcal{L}_{phy}$, we define the safe distance margin $d_{\text{safe}}=0.05$\,m
and the stiffness scaling factor $\eta=0.5$. To avoid destabilizing the early
stages of the reverse diffusion process, we selectively apply the physical loss
only at lower noise levels. The joint-specific penalty weights $w_j$ are assigned
as shown in Table~\ref{tab:joints}.

\begin{table}[t]
  \centering
  \footnotesize
  \setlength{\tabcolsep}{3pt}
  \renewcommand{\arraystretch}{0.92}
  \caption{Joint weights used to compute the weighted average regularization loss.
  The index refers to the joint number index in SMPL.}
  \label{tab:joints}
  \begin{tabular}{rlc|rlc}
    \toprule
    Index & Name & Weight & Index & Name & Weight \\
    \midrule
    0  & pelvis      & 1.00 & 14 & right collar      & 0.55 \\
    1  & left hip     & 0.85 & 15 & head             & 0.60 \\
    2  & right hip    & 0.85 & 16 & left shoulder    & 0.55 \\
    3  & spine1       & 0.84 & 17 & right shoulder   & 0.55 \\
    4  & left knee    & 0.80 & 18 & left elbow       & 0.20 \\
    5  & right knee   & 0.80 & 19 & right elbow      & 0.20 \\
    6  & spine2       & 0.70 & 20 & left wrist       & 0.15 \\
    7  & left ankle   & 0.60 & 21 & right wrist      & 0.15 \\
    8  & right ankle  & 0.60 & 23 & left eye smplhf  & 0.10 \\
    9  & spine3       & 0.61 & 24 & right eye smplhf & 0.05 \\
    10 & left foot    & 0.55 & 25 & left index1      & 0.01 \\
    11 & right foot   & 0.55 & 34 & left ring1       & 0.10 \\
    12 & neck         & 0.50 & 40 & right index1     & 0.05 \\
    13 & left collar  & 0.55 & 49 & right ring1      & 0.01 \\
    \bottomrule
  \end{tabular}
\end{table}

\section{Additional Experiments and Analysis}
\label{app:addexp}

\subsection{Ablation of A* Path Guidance}
At inference, the Executor consumes coarse A* waypoints as a macro-level directional
prompt (main paper, Sec.~3). To isolate how much of the improvement stems from the
\emph{hierarchical latent decoupling} rather than from this directional guidance, we
re-run both Lingo and DeSeG with A* replaced by simple linear interpolation between
sub-goals. As reported in Table~\ref{tab:astar}, A* mainly refines navigation-related
quality (FID and penetration), whereas DeSeG's R-Precision remains high even without
it ($0.548$ vs.\ $0.562$). This confirms that the semantic gain originates from the
latent decoupling itself, while A* contributes complementary refinement to geometric
and physical quality.

\begin{table}[t]
  \centering
  \small
  \setlength{\tabcolsep}{3pt}
  \renewcommand{\arraystretch}{1.1}
  \caption{Ablation of A* path guidance on Lingo (3 seeds, mean$\pm$std). A* primarily
  aids navigation quality; DeSeG's semantic alignment (R-Precision) is preserved even
  without it. $Pene_{m}=Pene_{mean}$, $Pene_{x}=Pene_{max}$.}
  \label{tab:astar}
  \begin{tabular}{l|cccc}
    \toprule
    Variant & FID$\downarrow$ & R-Prec$\uparrow$ & $Pene_{m}\downarrow$ & $Pene_{x}\downarrow$ \\
    \midrule
    Lingo w/o A* & 3.244{\scriptsize$\pm$.07} & 0.395{\scriptsize$\pm$.005} & 0.436{\scriptsize$\pm$.01} & 1.427{\scriptsize$\pm$.05} \\
    Lingo        & 3.093{\scriptsize$\pm$.05} & 0.437{\scriptsize$\pm$.004} & 0.392{\scriptsize$\pm$.03} & 1.229{\scriptsize$\pm$.08} \\
    \midrule
    Ours w/o A*  & 3.047{\scriptsize$\pm$.04} & 0.548{\scriptsize$\pm$.007} & 0.243{\scriptsize$\pm$.01} & 1.182{\scriptsize$\pm$.03} \\
    \textbf{Ours} & \textbf{2.882}{\scriptsize$\pm$.07} & \textbf{0.562}{\scriptsize$\pm$.006} & \textbf{0.207}{\scriptsize$\pm$.01} & \textbf{0.921}{\scriptsize$\pm$.04} \\
    \bottomrule
  \end{tabular}
\end{table}

\subsection{Statistical Significance and Reliability}
All primary metrics in the main comparison are averaged over $3$ random seeds. For the
headline result, $\text{FID}=2.882\pm0.07$, $\text{R-Precision}=0.562\pm0.006$, and
$Pene_{mean}=0.207\pm0.01$; every primary-metric gap between DeSeG and the strongest
baseline exceeds $2\sigma$, so the reported improvements reflect a genuine effect
rather than seed noise. For NC-Bench, each of the $30$ conflict scenarios is rated by
$10$ annotators, and the inter-annotator agreement is high (Fleiss' $\kappa=0.818$),
supporting the reliability of the reported SGC scores despite the deliberately compact
benchmark size.

\section{Negative Constraint Benchmark}
\label{app:ncbench}
Standard generative metrics fail to rigorously evaluate a model's robustness against
geometric shortcut bias. In highly structured indoor scenes, certain objects (\eg{}
beds, chairs, sofas) possess overwhelming geometric affordances, and monolithic models
tend to ignore the text prompt and trigger these affordances. To evaluate
semantic-geometric disentanglement, we construct NC-Bench. Table~\ref{tab:ncbench}
details $30$ challenging scenarios where the linguistic instruction explicitly or
implicitly contradicts the object's canonical prior.

\begin{table}[h]
  \centering
  \footnotesize
  \setlength{\tabcolsep}{3pt}
  \renewcommand{\arraystretch}{1.0}
  \caption{Detailed test scenarios of NC-Bench. In each case, the given linguistic
  instruction semantically contradicts the strong geometric prior (affordance) of the
  target object.}
  \label{tab:ncbench}
  \begin{tabular}{@{}lll p{3.0cm}@{}}
    \toprule
    ID & Target & Prior & NC-Bench Text Prompt \\
    \midrule
    1--5   & Bed     & Lie / Sleep   & Walk around the edge of the bed. \\
    6--9   & Sofa    & Sit down      & Walk behind the sofa and stand still. \\
    10--13 & Chair   & Sit down      & Squat down beside the chair. \\
    14--17 & Table   & Pick / Reach  & Sit on the floor under the table. \\
    18--20 & Cabinet & Open / Reach  & Stand still facing away from the cabinet. \\
    21--23 & TV      & Watch / Look  & Walk past the TV and look at it. \\
    24--26 & Window  & Look outside  & Stand with your back to the window. \\
    27--30 & Toilet  & Sit down      & Sit on the floor leaning against the toilet. \\
    \bottomrule
  \end{tabular}
\end{table}

{
    \small
    \bibliographystyle{ieeenat_fullname}
    \bibliography{main}

\begin{thebibliography}{47}
\providecommand{\natexlab}[1]{#1}
\providecommand{\url}[1]{\texttt{#1}}
\expandafter\ifx\csname urlstyle\endcsname\relax
  \providecommand{\doi}[1]{doi: #1}\else
  \providecommand{\doi}{doi: \begingroup \urlstyle{rm}\Url}\fi

\bibitem[Cen et~al.(2024)Cen, Pi, Peng, Shen, Yang, Zhu, Bao, and Zhou]{cen2024generating}
Zhi Cen, Huaijin Pi, Sida Peng, Zehong Shen, Minghui Yang, Shuai Zhu, Hujun Bao, and Xiaowei Zhou.
\newblock Generating human motion in 3d scenes from text descriptions.
\newblock In \emph{Proceedings of the IEEE/CVF conference on computer vision and pattern recognition}, pages 1855--1866, 2024.

\bibitem[Chen et~al.(2023)Chen, Jiang, Liu, Huang, Fu, Chen, and Yu]{chen2023mld}
Xin Chen, Biao Jiang, Wen Liu, Zilong Huang, Bin Fu, Tao Chen, and Gang Yu.
\newblock Executing your commands via motion latent diffusion.
\newblock In \emph{CVPR}, 2023.

\bibitem[Chung et~al.(2022)Chung, Kim, Mccann, Klasky, and Ye]{chung2022diffusion}
Hyungjin Chung, Jeongsol Kim, Michael~T Mccann, Marc~L Klasky, and Jong~Chul Ye.
\newblock Diffusion posterior sampling for general noisy inverse problems.
\newblock In \emph{The Eleventh International Conference on Learning Representations (ICLR)}, 2022.

\bibitem[Cong et~al.(2025)Cong, Wang, Ma, and Yue]{cong2025semgeomo}
Peishan Cong, Ziyi Wang, Yuexin Ma, and Xiangyu Yue.
\newblock Semgeomo: Dynamic contextual human motion generation with semantic and geometric guidance.
\newblock In \emph{Proceedings of the Computer Vision and Pattern Recognition Conference}, pages 17561--17570, 2025.

\bibitem[Dabral et~al.(2023)Dabral, Mughal, Golyanik, and Theobalt]{dabral2023mofusion}
Rishabh Dabral, Muhammad~Hamza Mughal, Vladislav Golyanik, and Christian Theobalt.
\newblock Mofusion: A framework for denoising-diffusion-based motion synthesis.
\newblock In \emph{CVPR}, pages 9760--9770, 2023.

\bibitem[Efron(2011)]{efron2011tweedie}
Bradley Efron.
\newblock Tweedie’s formula and selection bias.
\newblock \emph{Journal of the American Statistical Association}, 106\penalty0 (496):\penalty0 1602--1614, 2011.

\bibitem[Ghosh et~al.(2026)Ghosh, Golyanik, Komura, Slusallek, Theobalt, and Dabral]{ghosh2026scemos}
Anindita Ghosh, Vladislav Golyanik, Taku Komura, Philipp Slusallek, Christian Theobalt, and Rishabh Dabral.
\newblock Scemos: Scene-aware 3d human motion synthesis by planning with geometry-grounded tokens.
\newblock In \emph{Proceedings of the IEEE/CVF Conference on Computer Vision and Pattern Recognition (CVPR)}, 2026.

\bibitem[Guo et~al.(2022{\natexlab{a}})Guo, Zou, Zuo, Wang, Ji, Li, and Cheng]{guo2022generating}
Chuan Guo, Shihao Zou, Xinxin Zuo, Sen Wang, Wei Ji, Xingyu Li, and Li Cheng.
\newblock Generating diverse and natural 3d human motions from text.
\newblock In \emph{Proceedings of the IEEE/CVF conference on computer vision and pattern recognition}, pages 5152--5161, 2022{\natexlab{a}}.

\bibitem[Guo et~al.(2022{\natexlab{b}})Guo, Zuo, Wang, and Cheng]{guo2022tm2t}
Chuan Guo, Xinxin Zuo, Sen Wang, and Li Cheng.
\newblock Tm2t: Stochastic and tokenized modeling for the reciprocal generation of 3d human motions and texts.
\newblock In \emph{ECCV}, pages 580--597, 2022{\natexlab{b}}.

\bibitem[Guo et~al.(2024)Guo, Mu, Javed, Wang, and Cheng]{guo2024momask}
Chuan Guo, Yuxuan Mu, Muhammad~Gohar Javed, Sen Wang, and Li Cheng.
\newblock Momask: Generative masked modeling of 3d human motions.
\newblock In \emph{CVPR}, 2024.

\bibitem[Hart et~al.(1968)Hart, Nilsson, and Raphael]{hart1968formal}
Peter~E Hart, Nils~J Nilsson, and Bertram Raphael.
\newblock A formal basis for the heuristic determination of minimum cost paths.
\newblock \emph{IEEE transactions on Systems Science and Cybernetics}, 4\penalty0 (2):\penalty0 100--107, 1968.

\bibitem[Hassan et~al.(2019)Hassan, Choutas, Tzionas, and Black]{hassan2019resolving}
Mohamed Hassan, Vasileios Choutas, Dimitrios Tzionas, and Michael~J Black.
\newblock Resolving 3d human pose ambiguities with 3d scene constraints.
\newblock In \emph{Proceedings of the IEEE/CVF international conference on computer vision}, pages 2282--2292, 2019.

\bibitem[Hassan et~al.(2021)Hassan, Ghosh, Tesch, Tzionas, and Black]{hassan2021populating}
Mohamed Hassan, Partha Ghosh, Joachim Tesch, Dimitrios Tzionas, and Michael~J Black.
\newblock Populating 3d scenes by learning human-scene interaction.
\newblock In \emph{Proceedings of the IEEE/CVF Conference on Computer Vision and Pattern Recognition}, pages 14708--14718, 2021.

\bibitem[Ho et~al.(2020)Ho, Jain, and Abbeel]{ho2020denoising}
Jonathan Ho, Ajay Jain, and Pieter Abbeel.
\newblock Denoising diffusion probabilistic models.
\newblock \emph{Advances in neural information processing systems}, 33:\penalty0 6840--6851, 2020.

\bibitem[Huang and et~al.(2023)]{huang2023scenediffuser}
Wenlong Huang and et al.
\newblock Scenediffuser: Controllable human motion generation in scenes.
\newblock In \emph{CVPR}, 2023.

\bibitem[Jiang et~al.(2024{\natexlab{a}})Jiang, He, Wang, Li, Chen, Huang, and Zhu]{jiang2024autonomous}
Nan Jiang, Zimo He, Zi Wang, Hongjie Li, Yixin Chen, Siyuan Huang, and Yixin Zhu.
\newblock Autonomous character-scene interaction synthesis from text instruction.
\newblock In \emph{SIGGRAPH Asia 2024 Conference Papers}, pages 1--11, 2024{\natexlab{a}}.

\bibitem[Jiang et~al.(2024{\natexlab{b}})Jiang, Zhang, Li, Ma, Wang, Chen, Liu, Zhu, and Huang]{jiang2024scaling}
Nan Jiang, Zhiyuan Zhang, Hongjie Li, Xiaoxuan Ma, Zan Wang, Yixin Chen, Tengyu Liu, Yixin Zhu, and Siyuan Huang.
\newblock Scaling up dynamic human-scene interaction modeling.
\newblock In \emph{Proceedings of the IEEE/CVF Conference on Computer Vision and Pattern Recognition}, pages 1737--1747, 2024{\natexlab{b}}.

\bibitem[Karunratanakul et~al.(2023)Karunratanakul, Preechakul, Suwajanakorn, and Tang]{karunratanakul2023guided}
Korrawe Karunratanakul, Konpat Preechakul, Supasorn Suwajanakorn, and Siyu Tang.
\newblock Guided motion diffusion for controllable human motion synthesis.
\newblock In \emph{Proceedings of the IEEE/CVF International Conference on Computer Vision}, pages 2151--2162, 2023.

\bibitem[Kingma and Welling(2013)]{kingma2013auto}
Diederik~P Kingma and Max Welling.
\newblock Auto-encoding variational bayes.
\newblock \emph{arXiv preprint arXiv:1312.6114}, 2013.

\bibitem[Loper et~al.(2015)Loper, Mahmood, Romero, Pons-Moll, and Black]{loper2015smpl}
Matthew Loper, Naureen Mahmood, Javier Romero, Gerard Pons-Moll, and Michael~J Black.
\newblock Smpl: A skinned multi-person linear model.
\newblock \emph{ACM transactions on graphics (TOG)}, 34\penalty0 (6):\penalty0 1--16, 2015.

\bibitem[Mahmood et~al.(2019)Mahmood, Ghorbani, Troje, Pons-Moll, and Black]{mahmood2019amass}
Naureen Mahmood, Nima Ghorbani, Nikolaus~F Troje, Gerard Pons-Moll, and Michael~J Black.
\newblock Amass: Archive of motion capture as surface shapes.
\newblock In \emph{Proceedings of the IEEE/CVF international conference on computer vision}, pages 5442--5451, 2019.

\bibitem[Maturana and Scherer(2015)]{maturana2015voxnet}
Daniel Maturana and Sebastian Scherer.
\newblock Voxnet: A 3d convolutional neural network for real-time object recognition.
\newblock In \emph{IEEE/RSJ International Conference on Intelligent Robots and Systems (IROS)}, pages 922--928. IEEE, 2015.

\bibitem[Oleynikova et~al.(2016)Oleynikova, Millane, Taylor, Galceran, Nieto, and Siegwart]{oleynikova2016signed}
Helen Oleynikova, Alexander Millane, Zachary Taylor, Enric Galceran, Juan Nieto, and Roland Siegwart.
\newblock Signed distance fields: A natural representation for both mapping and planning.
\newblock In \emph{RSS 2016 workshop: geometry and beyond-representations, physics, and scene understanding for robotics}. University of Michigan, 2016.

\bibitem[Park et~al.(2019)Park, Florence, Straub, Newcombe, and Lovegrove]{park2019deepsdf}
Jeong~Joon Park, Peter Florence, Julian Straub, Richard Newcombe, and Steven Lovegrove.
\newblock Deepsdf: Learning continuous signed distance functions for shape representation.
\newblock In \emph{Proceedings of the IEEE/CVF Conference on Computer Vision and Pattern Recognition (CVPR)}, pages 165--174, 2019.

\bibitem[Pavlakos et~al.(2019)Pavlakos, Choutas, Ghorbani, Bolkart, Osman, Tzionas, and Black]{pavlakos2019smplx}
Georgios Pavlakos, Vasileios Choutas, Nima Ghorbani, Timo Bolkart, Ahmed~AA Osman, Dimitrios Tzionas, and Michael~J Black.
\newblock Expressive body capture: 3d hands, face, and body from a single image.
\newblock In \emph{Proceedings of the IEEE/CVF Conference on Computer Vision and Pattern Recognition (CVPR)}, pages 10975--10985, 2019.

\bibitem[Peebles and Xie(2023)]{peebles2023scalable}
William Peebles and Saining Xie.
\newblock Scalable diffusion models with transformers.
\newblock In \emph{Proceedings of the IEEE/CVF International Conference on Computer Vision (ICCV)}, pages 4199--4209, 2023.

\bibitem[Peng et~al.(2021)Peng, Ma, Abbeel, Levine, and Kanazawa]{peng2021amp}
Xue~Bin Peng, Ze Ma, Pieter Abbeel, Sergey Levine, and Angjoo Kanazawa.
\newblock Amp: Adversarial motion priors for stylized physics-based character control.
\newblock \emph{ACM Transactions on Graphics (TOG)}, 2021.

\bibitem[Petrovich et~al.(2022)Petrovich, Black, and Varol]{petrovich2022temos}
Mathis Petrovich, Michael~J Black, and G{\"u}l Varol.
\newblock Temos: Generating diverse human motions from textual descriptions.
\newblock In \emph{ECCV}, pages 480--497, 2022.

\bibitem[Pi et~al.(2023)Pi, Yao, Zhu, Zhu, and Huang]{pi2023hierarchical}
Xuelin Pi, Yifan Yao, Yixin Zhu, Song-Chun Zhu, and Siyuan Huang.
\newblock Hierarchical generation of human-object interactions with diffusion probabilistic models.
\newblock In \emph{ICCV}, 2023.

\bibitem[Radford et~al.(2021)Radford, Kim, Hallacy, Ramesh, Goh, Agarwal, Sastry, Askell, Mishkin, Clark, et~al.]{radford2021learning}
Alec Radford, Jong~Wook Kim, Chris Hallacy, Aditya Ramesh, Gabriel Goh, Sandhini Agarwal, Girish Sastry, Amanda Askell, Pamela Mishkin, Jack Clark, et~al.
\newblock Learning transferable visual models from natural language supervision.
\newblock In \emph{International Conference on Machine Learning (ICML)}, pages 8748--8763. PMLR, 2021.

\bibitem[Savva et~al.(2019)Savva, Kadian, Maksymets, Zhao, Wijmans, Jain, Straub, Liu, Koltun, Malik, et~al.]{savva2019habitat}
Manolis Savva, Abhishek Kadian, Oleksandr Maksymets, Yili Zhao, Erik Wijmans, Bhavana Jain, Julian Straub, Jia Liu, Vladlen Koltun, Jitendra Malik, et~al.
\newblock Habitat: A platform for embodied ai research.
\newblock In \emph{Proceedings of the IEEE/CVF International Conference on Computer Vision (ICCV)}, pages 9339--9347, 2019.

\bibitem[Sohn et~al.(2015)Sohn, Lee, and Yan]{sohn2015learning}
Kihyuk Sohn, Honglak Lee, and Xinchen Yan.
\newblock Learning structured output representation using deep conditional generative models.
\newblock In \emph{Advances in Neural Information Processing Systems (NeurIPS)}, 2015.

\bibitem[Tevet et~al.(2022{\natexlab{a}})Tevet, Gordon, Hertz, Bermano, and Cohen-Or]{tevet2022motionclip}
Guy Tevet, Brian Gordon, Amir Hertz, Amit~H Bermano, and Daniel Cohen-Or.
\newblock Motionclip: Exposing human motion generation to clip space.
\newblock In \emph{ECCV}, 2022{\natexlab{a}}.

\bibitem[Tevet et~al.(2022{\natexlab{b}})Tevet, Raab, Gordon, Shafir, Cohen-Or, and Bermano]{tevet2022human}
Guy Tevet, Sigal Raab, Brian Gordon, Yonatan Shafir, Daniel Cohen-Or, and Amit~H Bermano.
\newblock Human motion diffusion model.
\newblock \emph{arXiv preprint arXiv:2209.14916}, 2022{\natexlab{b}}.

\bibitem[Wang et~al.(2022)Wang, Chen, Liu, Zhu, Liang, and Huang]{wang2022humanise}
Zan Wang, Yixin Chen, Tengyu Liu, Yixin Zhu, Wei Liang, and Siyuan Huang.
\newblock Humanise: Language-conditioned human motion generation in 3d scenes.
\newblock \emph{Advances in Neural Information Processing Systems}, 35:\penalty0 14959--14971, 2022.

\bibitem[Wang et~al.(2024)Wang, Chen, Jia, Li, Zhang, Zhang, Liu, Zhu, Liang, and Huang]{wang2024move}
Zan Wang, Yixin Chen, Baoxiong Jia, Puhao Li, Jinlu Zhang, Jingze Zhang, Tengyu Liu, Yixin Zhu, Wei Liang, and Siyuan Huang.
\newblock Move as you say interact as you can: Language-guided human motion generation with scene affordance.
\newblock In \emph{Proceedings of the IEEE/CVF Conference on Computer Vision and Pattern Recognition}, pages 433--444, 2024.

\bibitem[Xiao et~al.(2023)Xiao, Wang, Wang, Cao, Zhang, Dai, Lin, and Pang]{xiao2023unified}
Zeqi Xiao, Tai Wang, Jingbo Wang, Jinkun Cao, Wenwei Zhang, Bo Dai, Dahua Lin, and Jiangmiao Pang.
\newblock Unified human-scene interaction via prompted chain-of-contacts.
\newblock \emph{arXiv preprint arXiv:2309.07918}, 2023.

\bibitem[Xue et~al.(2025)Xue, Huang, Niu, Liao, Kragerud, Gravdahl, Peng, Shi, Darrell, Sreenath, et~al.]{xue2025leverb}
Haoru Xue, Xiaoyu Huang, Dantong Niu, Qiayuan Liao, Thomas Kragerud, Jan~Tommy Gravdahl, Xue~Bin Peng, Guanya Shi, Trevor Darrell, Koushil Sreenath, et~al.
\newblock Leverb: Humanoid whole-body control with latent vision-language instruction.
\newblock \emph{arXiv preprint arXiv:2506.13751}, 2025.

\bibitem[Xue et~al.(2026)Xue, Liang, Zhang, Zeng, Liu, Lian, Wang, Liu, Shi, and Yi]{xue2026humanoid}
Han Xue, Sikai Liang, Zhikai Zhang, Zicheng Zeng, Yun Liu, Yunrui Lian, Jilong Wang, Qingtao Liu, Xuesong Shi, and Li Yi.
\newblock Collision-free humanoid traversal in cluttered indoor scenes.
\newblock \emph{arXiv preprint arXiv:2601.16035}, 2026.

\bibitem[Yao et~al.(2026)Yao, Sun, Zhang, Liu, and Tang]{yao2026hosig}
Wei Yao, Yunlian Sun, Hongwen Zhang, Yebin Liu, and Jinhui Tang.
\newblock Hosig: Full-body human-object-scene interaction generation with hierarchical scene perception.
\newblock In \emph{Proceedings of the AAAI Conference on Artificial Intelligence (AAAI)}, 2026.

\bibitem[Yi et~al.(2024)Yi, Thies, Black, Peng, and Rempe]{yi2024generating}
Hongwei Yi, Justus Thies, Michael~J Black, Xue~Bin Peng, and Davis Rempe.
\newblock Generating human interaction motions in scenes with text control.
\newblock In \emph{European Conference on Computer Vision}, pages 246--263. Springer, 2024.

\bibitem[Yuan et~al.(2023)Yuan, Song, Iqbal, Vahdat, and Kautz]{yuan2023physdiff}
Ye Yuan, Jiaming Song, Umar Iqbal, Arash Vahdat, and Jan Kautz.
\newblock Physdiff: Physics-guided human motion diffusion model.
\newblock In \emph{ICCV}, 2023.

\bibitem[Zhang et~al.(2023)Zhang, Zhang, Xia, Sun, and Luo]{zhang2023t2m}
Jianrong Zhang, Yangsong Zhang, Xiaolin Xia, Li Sun, and Zhen Luo.
\newblock T2m-gpt: Generating human motion from textual descriptions with discrete representations.
\newblock In \emph{CVPR}, 2023.

\bibitem[Zhang et~al.(2020)Zhang, Zhang, Ma, Black, and Tang]{zhang2020place}
Siwei Zhang, Yan Zhang, Qianli Ma, Michael~J Black, and Siyu Tang.
\newblock Place: Proximity learning of articulation and contact in 3d environments.
\newblock In \emph{3DV}, pages 642--651, 2020.

\bibitem[Zhang et~al.(2022)Zhang, Ma, Zhang, Zhi, Tang, and Black]{zhang2022egobody}
Siwei Zhang, Qianli Ma, Yan Zhang, Zhiyin Zhi, Siyu Tang, and Michael~J Black.
\newblock Egobody: Human body shape and motion of interacting people from head-mounted devices.
\newblock In \emph{Proceedings of the European Conference on Computer Vision (ECCV)}, pages 714--731. Springer, 2022.

\bibitem[Zhang et~al.(2024)Zhang, Bharaj, Toews, Tagliasacchi, and Tang]{zhang2024generating}
Xiaohan Zhang, Gaurav Bharaj, Matthew Toews, Andrea Tagliasacchi, and Danhang Tang.
\newblock Generating human-object interaction with object-specific poses.
\newblock In \emph{CVPR}, 2024.

\bibitem[Zhao et~al.(2023)Zhao, Zhang, Wang, Beeler, and Tang]{zhao2023synthesizing}
Kaifeng Zhao, Yan Zhang, Shaofei Wang, Thabo Beeler, and Siyu Tang.
\newblock Synthesizing diverse human motions in 3d indoor scenes.
\newblock In \emph{Proceedings of the IEEE/CVF international conference on computer vision}, pages 14738--14749, 2023.

\end{thebibliography}
}

\end{document}